\documentclass[10pt,twocolumn,letterpaper]{article}

\usepackage[font=footnotesize]{caption}
\usepackage{cvpr}
\usepackage{times}
\usepackage{epsfig}
\usepackage{graphicx}
\usepackage{amsmath}
\usepackage{amssymb}
\usepackage{breqn}
\usepackage{subfig}
\usepackage{multicol}
\usepackage{float}
\usepackage{lipsum}
\usepackage{authblk}
\usepackage{subfig}
\usepackage{caption}
\usepackage{pifont}
\newcommand{\cmark}{\ding{51}}%
\newcommand{\xmark}{\ding{55}}%

\DeclareMathOperator{\argmax}{argmax}

\usepackage[pagebackref=true,breaklinks=true,letterpaper=true,colorlinks,bookmarks=false]{hyperref}

\cvprfinalcopy 

\def\cvprPaperID{****} 

\ifcvprfinal\fi
\begin{document}

\newcommand{\rlc}{RL-CycleGAN }
\newcommand{\sr}{simulation-to-real }
\newcommand{\rlcnospace}{RL-CycleGAN}
\newcommand{\srnospace}{simulation-to-real}
\newcommand{\kuka}{Robot 1}
\newcommand{\sorty}{Robot 2 }

\title{RL-CycleGAN: Reinforcement Learning Aware Simulation-To-Real}

\author[1]{Kanishka Rao}
\author[1]{Chris Harris}
\author[1]{Alex Irpan}
\author[1, 2]{Sergey Levine}
\author[1]{Julian Ibarz}
\author[3]{Mohi Khansari}
\affil[1]{Google Brain, Mountain View, USA}
\affil[2]{University of California Berkeley, Berkeley, USA}
\affil[3]{X, The Moonshot Factory, Mountain View, USA}
\affil[ ]{\{kanishkarao, ckharris, alexirpan, slevine, julianibarz\}@google.com,  khansari@x.team}


\twocolumn[{%
\renewcommand\twocolumn[1][]{#1}%
\vspace{-3em}
\maketitle
\vspace{-3em}
\begin{center}
    \label{robot2}
    \includegraphics[width=0.9\linewidth]{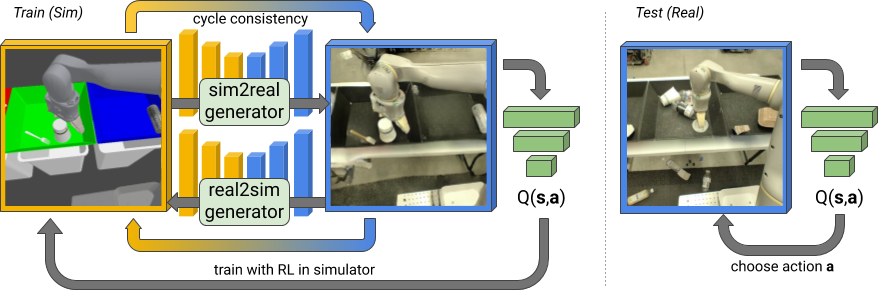}
    \captionof{figure}{\footnotesize \rlc trains a CycleGAN which maps an image from the simulator (left) to a realistic image (middle), a jointly trained RL task ensures that these images are useful for that specific task. At test time, the RL model may be transferred to real robot (right).}
    \vspace{1em}
\end{center}%
}]

\begin{abstract}
\vspace{-3mm}
Deep neural network based reinforcement learning (RL) can learn appropriate visual representations for complex tasks like vision-based robotic grasping without the need for manually engineering or prior learning a perception system. However, data for RL is collected via running an agent in the desired environment, and for applications like robotics, running a robot in the real world may be extremely costly and time consuming. Simulated training offers an appealing alternative, but ensuring that policies trained in simulation can transfer effectively into the real world requires additional machinery. Simulations may not match reality, and typically bridging the simulation-to-reality gap requires domain knowledge and task-specific engineering. We can automate this process by employing generative models to translate simulated images into realistic ones. However, this sort of translation is typically task-agnostic, in that the translated images may not preserve all features that are relevant to the task. In this paper, we introduce the RL-scene consistency loss for image translation, which ensures that the translation operation is invariant with respect to the Q-values associated with the image. This allows us to learn a task-aware translation. Incorporating this loss into unsupervised domain translation, we obtain \rlcnospace, a new approach for simulation-to-real-world transfer for reinforcement learning. In evaluations of \rlc on two vision-based robotics grasping tasks, we show that \rlc offers a substantial improvement over a number of prior methods for sim-to-real transfer, attaining excellent real-world performance with only a modest number of real-world observations.
\end{abstract}

\vspace{-5mm}
\section{Introduction}
\label{intro}


Reinforcement learning (RL) can be used to train deep neural network models to grasp objects directly with image observations~\cite{zeng2018,learning_by_play}, or perform navigation with a mobile robot directly from onboard sensor readings~\cite{long_range_nav}. However, this ability to learn visual representations end-to-end together with a task controller often comes at a steep price in sample complexity. Since the data needed for RL is typically task and policy specific, collecting this data in the loop with policy training can be particularly difficult. An appealing alternative is to use RL to train policies in simulation, and then transfer these policies onto real-world systems. For acquiring task-relevant visual representations, training in simulation is suboptimal as it results in representations of the \emph{simulated} environment, which may not work as well for \emph{real} environments. This simulation-to-reality \emph{gap} has been addressed in a variety of ways in prior work, from employing domain adaptation techniques that modify the simulated training images automatically~\cite{Bousmalis2017UsingSA} to randomizing the simulation environment in the hopes that such randomization will improve the transferability of the learned representations~\cite{Sadeghi2016CAD2RLRS, Tobin2017DomainRF, James2017TransferringEV, Matas2018SimtoRealRL, James2018TaskEmbeddedCN}. However, the objective function of these approaches are generally \emph{task-agnostic}. This often requires having to adapt these methods to each individual task through manual modification. Traditionally researchers have either increased their diversity (such as domain adaptation) or directly modified their methods to appear more realistic (such as pixel-level domain adaptation).

%
We propose a method to automatically transfer vision-based policies from simulation with an objective that is \emph{task-aware}, but still automated, in the sense that it does not require task-specific engineering. To avoid the manual engineering required to produce randomized simulation environments, we automatically translate simulated observations into realistic ones via a generative adversarial network (GAN). We assume access to an \emph{off-policy} dataset of real experience, which would typically be collected either randomly or with a low-performing exploration policy, and we do not assume access to paired simulated data. We employ a cycle consistency approach for training this model, following the CycleGAN method~\cite{Zhu2017UnpairedIT}. This provides pixel-level domain adaptation for the simulated images, allowing us to train in simulation on images that resemble those that the policy would see in the real world. Enforcing cycle consistency during GAN training encourages the adapted image to retain certain attributes of the input image, since it must be reconstructed. However, which attributes are retained is not enforced. To be useful for RL, it is extremely important that the GAN adaptation retains all the attributes that might affect the RL outcome. For example, in robotic grasping the GAN may alter the lighting and object textures, but must not change the location of the robot arm or objects. In the case of grasping we may construct additional losses that preserve the scene geometry~\cite{Bousmalis2017UsingSA}, however, this solution is task-specific. To address this challenge in a task-independent way, we introduce the \emph{RL-scene consistency loss}, which enforces that the Q-values predicted by an RL-trained Q-function should be invariant under the CycleGAN transformation. This loss is general, in that it can be utilized for any reinforcement learning problem, and we find empirically that our proposed \emph{RL-CycleGAN} substantially improves transfer performance over a standard CycleGAN that is \emph{task-agnostic}.

%
Vision-based tasks are particularly suitable for testing visual simulation-to-real methods but may not address physics-based simulation-to-real gap due to poorly simulated dynamics. Our method, which adapts a single state (an image in this case), does not address the physics gap. We investigate simulation-to-real for vision-based grasping tasks with two different robotic systems that are both learned with a reinforcement learning method, QT-Opt~\cite{Kalashnikov2018ScalableDR}. In RL, real-world episodes are considered {\it off-policy} if they are collected with a scripted policy or a previously trained model. Episodes are considered {\it on-policy} if they are collected with the latest policy. Training on off-policy episodes is significantly more practical, as the same data can be reused across different training runs and no new real robot episodes are necessary. Therefore, it is highly desirable to have a learning system that does not require any \emph{on-policy} real-world trials, as such a system could be trained entirely from simulated data and logged real data, without any additional real-world data collection during a training run.
We primarily experiment in the scenario where only off-policy real data is available, but also provide comparisons for how \rlc may be used with on-policy real-world training.

\paragraph{Contributions} We introduce \rlcnospace, which enables {\it RL-aware} simulation-to-real with a CycleGAN constrained by an \emph{RL-scene consistency loss} for vision-based reinforcement learning policies. With our approach, the CycleGAN losses encourage some preservation of the input image, while the RL-scene consistency loss specifically focuses on those features that are most critical for the current RL-trained Q-function. We show how our \emph{RL-aware} simulation-to-real can be used to train policies with simulated data, utilizing only domain adaptation techniques that modify mph{off-policy} real data. \rlc does not require per-task manual engineering, unlike several related methods that utilize randomization or task-specific losses. We demonstrate our approach on two real-world robotic grasping tasks, showing that RL-CycleGAN achieves efficient transfer with very high final performance in the real world, and substantially outperforms a range of prior approaches.
\section{Related Work}
\label{related}



It is relatively easy to generate a large amount of simulation data with oracle labels, which makes model development in simulation especially attractive. However, such models tend to perform poorly when evaluated on real robots since
the simulated data may differ from the real world both visually and physically. We focus on the visual simulation-to-real gap where the simulated images may have unrealistic textures, lighting, colors, or objects.
To address the visual simulation-to-real gap, various recent works use randomized simulated environments~\cite{Tobin2017DomainRF, Matas2018SimtoRealRL, James2017TransferringEV, Sadeghi2017}
to randomize the textures, lighting, cropping and camera position. These models are more robust when transferred to a real robot, since they train on diverse data and the real world may be within the distribution of randomization used. However, such randomization requires manually defining what aspects of the simulator to randomize. For example, with grasping, if it is observed that the simulated object textures differ from those in the real world, applying randomization to those textures may lead to texture-robust models with improved real world performance. Our proposed approach does not require  manually instrumenting the simulator, and can be seen as a visual domain adaptation technique~\cite{Patel2015VisualDA} that learns directly from a data set of real images. Domain adaptation methods aim to train models using many examples from a source domain (simulation) and few examples from a target domain (reality). Prior methods can be split into \textit{feature-level} adaptation, where they learn domain-invariant features~\cite{gopalan2011domain,caseiro2015beyond,long2015learning,ganin2016domain}, and \textit{pixel-level} adaptation, where they condition on pixels from a source image and re-style it look like an image from the target domain~\cite{Bousmalis2016UnsupervisedPD, Yoo2016PixelLevelDT, Hua2017UnsupervisedCI, Hoffman2017}.

Pixel-level adaptation is an especially challenging image-translation task when we do not have paired data. Prior techniques tackle this problem using generative adversarial networks (GANs)~\cite{Goodfellow2014GenerativeAN,Zhang2018SelfAttentionGA, Brock2018LargeSG}, conditioning the GAN generator on the simulated image's pixels. Our technique is based on the CycleGAN pixel-level adaptation approach, with additional RL specific losses.

One related pixel-level method is RCAN~\cite{James2018SimtoRealVS}, which learns a model mapping images from randomized simulations to a canonical simulation.
Robotic grasping models are trained on canonical simulated images from the RCAN generator, and at inference time the generator maps real images to the canonical simulator. This approach still requires manually defining the task-specific canonical scene components and the corresponding simulator randomization.
Real-to-simulation methods like RCAN also require adapting
real-world images at inference time,
which can be computationally prohibitive when the RCAN generator has many more parameters than the task model.

A central challenge in using a GAN for simulation-to-real transfer is that, by design, a GAN learns to generate any image from the real distribution which may not correspond to the input simulated image. For simulation-to-real we want a {\it realistic version} of the input simulated image, not just any realistic image.
GraspGAN~\cite{Bousmalis2017UsingSA} addresses this for robotic grasping by having the GAN reproduce the segmentation mask for the simulated image as an auxiliary task, which includes the robot arm, objects, and the bin. GraspGAN further constrains the GAN by enforcing a feature-level domain-adversarial loss. We show that RL and CycleGAN consistency losses let us outperform GraspGAN without using task-specific semantic segmentation or feature-level domain adaptation.

Recently, the CycleGAN~\cite{Zhu2017UnpairedIT} was proposed for unpaired image-to-image translation between domains. This involves two GANs, one to adapt from the source to the target domain and the other to adapt from the target to the source. A cycle consistency loss ensures that the GANs applied in succession recreates the original image, which encourages preserving aspects of the original image since they must be reproduced.
This is especially attractive for the simulation-to-real gap, where we want to adapt visual differences but retain semantics relevant to the RL task.
However, CycleGANs may learn to hide information in the adapted image instead of explicitly retaining the semantics~\cite{Chu2017CycleGANAM}, or may change them in a deterministic way that is reversed by the other generator. We mitigate these undesirable CycleGAN behaviors by jointly training an RL model that informs the GAN about which components of the image are relevant for RL by enforcing RL consistency losses on all the input and generated images.


We evaluate our method on robotic grasping. Grasping is one of the most fundamental robotics problems and has yielded a large variety of research. A thorough survey can be found in~\cite{Bohg2013DataDrivenGS}.
Recent state-of-the-art results have come from deep-learning based methods~\cite{Lenz2013DeepLF, dexnet} that make use of
hand-labeled grasp positions or predicting grasp outcomes in an RL setup. In this work, we consider closed-loop grasping
where grasp prediction is continuously made during prediction. We consider the vision-based grasping model as described in~\cite{Kalashnikov2018ScalableDR}, via Q-learning with a deep neural network conditioned on an RGB image and the proposed action. 

\section{Preliminaries}
Our approach is based off of combining CycleGAN with a Q-learning task model. We briefly cover both of those techniques.
\subsection{CycleGAN}
CycleGANs are a technique for learning a mapping between two image domains $X$ and $Y$, from unpaired examples $\{x_i\}^N_{i=1} \in X$ and $\{y_i\}^M_{i=1} \in Y$. For \srnospace, $X$ and $Y$ are simulation and real respectively. Following the notation in~\cite{Zhu2017UnpairedIT}, the CycleGAN involves learning two generators: Sim2Real, ${G : X \rightarrow Y}$ and Real2Sim,  ${F : Y \rightarrow X}$. Two adversarial discriminators $D_X$ and $D_Y$ distinguish simulated images $\{x\}$ from adapted simulation $\{F(y)\}$ and real images $\{y\}$ from adapted real $\{G(x)\}$.

An {\it adversarial loss} is applied to both mappings. For Sim2Real, the loss is:
\begin{dmath}
\mathcal{L}_{GAN}(G, D_Y, X, Y) = \mathbb{E}_{y\sim Y}[\log D_Y(y)] 
                                    + \mathbb{E}_{x\sim X}[\log(1 - D_Y(G(x)))]
\end{dmath}

The Sim2Real generator $G$ aims to produce realistic images by minimizing this objective against an adversary $D_Y$ that tries to maximize it, giving update $\min_G \max_{D_Y}\mathcal{L}_{GAN}(G, D_Y, X, Y)$. Real2Sim is trained similarly, with $\min_F \max_{D_X}\mathcal{L}_{GAN}(F, D_X, Y, X)$.
The CycleGAN further imposes a {\it cycle consistency loss}, to encourage ${x \to G(x) \to F(G(x)) \approx x}$ and ${y \to F(y) \to G(F(y)) \approx y}$.

\begin{dmath}
\mathcal{L}_{cyc}(G, F) =  \mathbb{E}_{x\sim\mathcal{D}_{sim}}d(F(G(x)), x) + \mathbb{E}_{y\sim\mathcal{D}_{real}}d(G(F(y)), y)
\end{dmath}

Here, $d$ is some distance metric. We use mean squared error. This cycle-consistency prevents drastic departures in the generated, as the original scene must be recoverable, but~\cite{Chu2017CycleGANAM} argues the scene may still be altered even with this consistency loss.

\subsection{Q-learning}
Given an environment of states $\{s\}$, actions $\{a\}$, rewards $\{r\}$, and next states $\{s'\}$,
Q-learning is a reinforcement learning technique that learns a Q-function $Q(s,a)$, representing total expected future reward~\cite{watkins1992q}. For a vision-based task, $s$ is the input image, and $a$ a candidate action. The Q-function is updated to minimize the temporal difference (TD) loss, defined as

\begin{equation}
    d(Q(s,a), r + \gamma V(s'))
\end{equation}

\noindent where $V(s')$ is an estimate of the next state's value, $\gamma$ is a discount factor, and $d$ is a distance metric. The policy $\pi(a|s)$ is then defined by $\argmax_a Q(s,a)$. To estimate $V(s')$, we use Clipped Double-Q Learning~\cite{fujimoto2018addressing}. RL-CycleGAN jointly trains a Q-function with the CycleGAN, using the learned Q-values to add additional consistency losses.
\section{\rlc}
\label{rlcgan}

\begin{figure*}
\begin{center}
\includegraphics[scale=0.6]{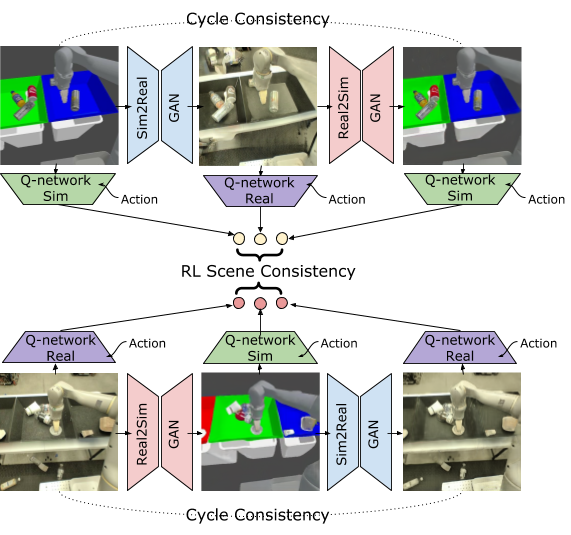}
\end{center}
\vspace{-5mm}
\caption{\footnotesize \rlc involves a CycleGAN constrained with an RL scene consistency. A simulated state for grasping (top-left image) is adapted by a Sim2Real GAN to be more realistic (top-center), a further cycled adaptation by a Real2Sim GAN (top-right) is required to match the original simulated input. A sim Q-network is trained on the original (top-left) and cycled (top-right) simulated images along with the simulated action via a TD-loss. Another real Q-network is trained with the realistic image (top-center) and simulated action. Finally, the GAN generators are constrained with RL-scene consistency which requires the same Q-values are produced for all three images. In the bottom row, the same neural networks are required to satisfy the same constraints with a real image and corresponding action.}
\label{fig:rl-cyclegan-diagram}
\vspace{-2mm}
\end{figure*}

The key for a useful \sr model is to adapt simulated images to realistic images while also preserving the original semantics relevant to the RL task. For example with grasping, a \sr model may produce very realistic images, but in the process may remove some of the objects from the image if they are not easily transformed in to realistic versions. Such alterations drastically change the grasping outcome and are detrimental to the RL task. The distinction between {\it style} (lighting, textures, etc) that does not affect the task and {\it semantics} (robot and object positions) that does affect the task is not always clear and varies with the task. We introduce \rlc, which trains a GAN that is encouraged to make this {\it style-semantics} distinction via a jointly trained RL model. Intuitively, the RL model's output should only depend on the {\it semantics} of the task, and constraining the GAN with the RL model encourages the GAN to preserve task-specific semantics.

The RL task model is a deep Q-learning network $Q(s, a)$. For a vision-based task, $s$ is the input image and $a$ a candidate action. $Q_{sim}(s,a)$ and $Q_{real}(s,a)$ represent Q-functions trained on simulated and real $(s,a)$ respectively. The \rlc jointly trains the RL model with the CycleGAN, where each of the 6 images $\{x, G(x), F(G(x))\}$ and $\{y, F(y), G(F(y)\}$ are passed to $Q_{sim}$ and $Q_{real}$, giving 6 Q-values.
\begin{gather*}
(x,a) \sim \mathcal{D}_{sim}, (y,a) \sim \mathcal{D}_{real} \\
q_x =  Q_{sim}(x, a)  \\
q_x^\prime =  Q_{real}(G(x), a) \\
q_{x}^{\prime\prime} =  Q_{sim}(F(G(x)), a) \\
q_y =  Q_{real}(y, a)  \\
q_y^\prime =  Q_{sim}(F(y), a) \\
q_{y}^{\prime\prime} =  Q_{real}(G(F(y)), a) 
\end{gather*}

These $q$ represent the Q-values for the various images. Triples $\{x, G(x), F(G(x))\}$ and $\{y, F(y), G(F(y))\}$ should each represent the same scene, and an {\it RL-scene consistency loss} is imposed by encouraging similar Q-values within the triple.

\begin{dmath}
\mathcal{L}_{RL-scene}(G, F) =  d(q_x, q_x^\prime) + d(q_x, q_{x}^{\prime\prime}) + d(q_x^\prime, q_{x}^{\prime\prime}) + d(q_y, q_y^\prime) + d(q_y, q_{y}^{\prime\prime}) + d(q_y^\prime, q_{y}^{\prime\prime})
\end{dmath}
Again, $d$ is some distance metric, and we use mean squared error. This loss penalizes changes in the Q-value, further encouraging preserving the RL-scene during adaptation. Since visual features for grasping in simulation and reality might differ drastically, we train two different Q-networks $Q_{sim}, Q_{real}$ to compute Q-values for simulation-like and real-like images.
These Q-networks are trained via the standard TD-loss, on all original and generated images $\{x,F(G(x)), F(y)\}$ for $Q_{sim}$ and $\{G(x),y,G(F(y))\}$ for $Q_{real}$. Each generator or pair of generators is applied to both current image $x$ and next image $x'$ before the TD-loss is computed.
\begin{dmath}
\mathcal{L}_{RL}(Q) =  \mathbb{E}_{(x,a,r,x')}d(Q(x, a), r + \gamma V(x'))
\end{dmath}
The full objective is:

\begin{dmath}
\mathcal{L}_{RL-CycleGAN}(G, F, D_X, D_Y, Q) = \lambda_{GAN}\mathcal{L}_{GAN}(G, D_Y) + \lambda_{GAN}\mathcal{L}_{GAN}(F, D_X) + \lambda_{cycle}\mathcal{L}_{cyc}(G, F) + \lambda_{RL-scence}\mathcal{L}_{RL-scene}(G, F) + \lambda_{RL}\mathcal{L}_{RL}(Q)
\end{dmath}
where the $ \lambda $ are relative loss weights.

A diagram of \rlc is shown in Figure~\ref{fig:rl-cyclegan-diagram}. All \rlc neural networks are trained jointly from scratch using the distributed Q-learning QT-Opt algorithm. Simulated $(s, a)$ are generated from a simulator and real $(s, a)$ are read from off-policy episodes. After the RL-CycleGAN is learned, the $Q_{real}$ learned could be used for the final real-world policy, but we found we got best performance by freezing the Sim2Real generator and retraining a $Q(s,a)$ from scratch.
\section{Task Setup}

We evaluate our methods on two real world robot grasping setups, which use different physical robots, objects, bins, and simulation environments. Robot 1's setup aims to generalize grasping of unseen objects, while Robot 2 grasps from three bins with the robot placed at different locations relative to the bin. We aim to show our approach is independent of robot and task, and do not tailor the \rlc for either setup. Both tasks perform dynamic closed-loop grasping~\cite{Bohg2013DataDrivenGS, Dafle2018} with sensing and control tightly interleaved at each stage and trained as described in ~\cite{Kalashnikov2018ScalableDR}. Observations consist of monocular RGB camera images. Actions directly command the robot gripper in four dimensions (xyz and top-down rotation), along with gripper close/open and episode termination commands. 

\subsection{Robot 1 Setup}

\label{robots}
\begin{figure*}
\begin{center}
\includegraphics[width=\textwidth]{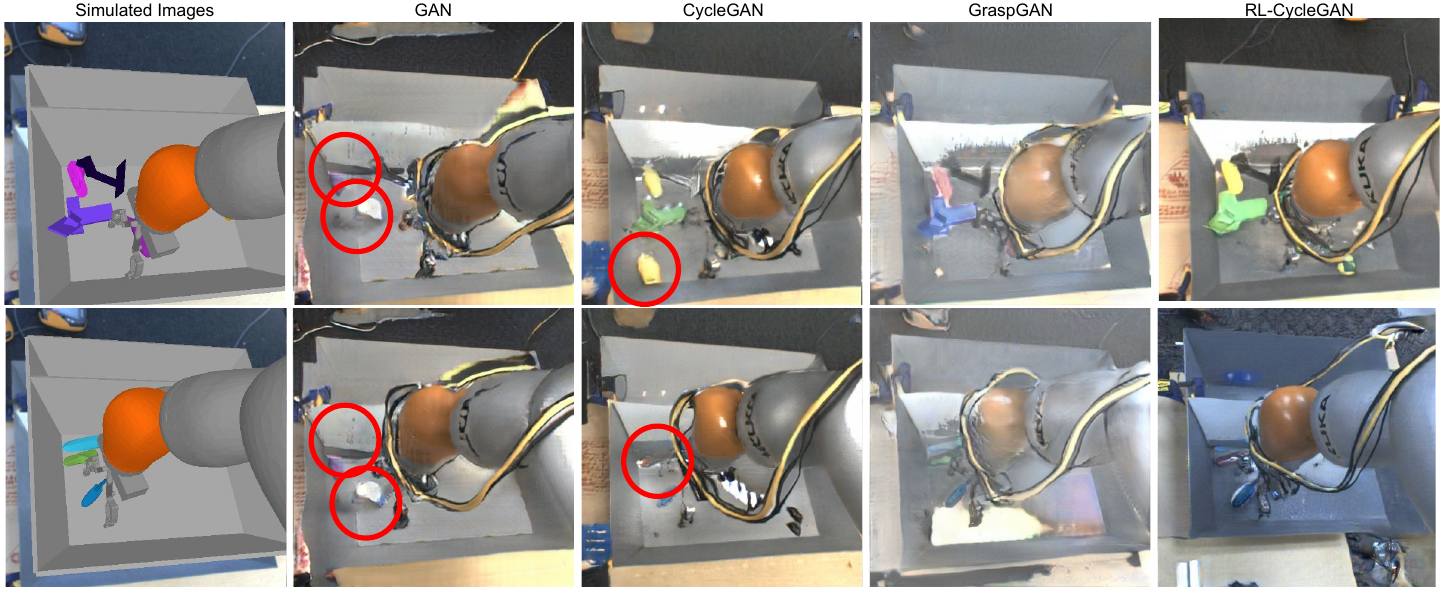}
\end{center}
\vspace{-4mm}
\caption{\footnotesize Examples of simulation-to-real for four models shown on two simulated images (left). The GAN alone produces poor images with objects deleted or added and unclear robot gripper locations (shown with red circles). CycleGAN is better at reproducing objects and the robot, however, some object deletion is still seen. GraspGAN retains the objects but overall image quality is poor, especially for the robot arm and gripper. Although, some artifacts are still seen, \rlc produces the best images, retaining all the RL task information while producing realistic images. An interesting conflict is seen in some examples where the objects are generated on top of the the generated robot wires.}
\label{fig:kuka-gans}
\vspace{-2mm}
\end{figure*}

We use Kuka IIWA robots to grasp a variety of objects from a metal bin as in~\cite{Kalashnikov2018ScalableDR}.
Real robot episodes are collected by running a scripted policy or a previously learned model using training objects.
A simulated environment for the task is also built using the Bullet physics simulator~\cite{coumans2015bullet}, containing the robot arm, the bin, and the objects to be grasped. In order to generalize grasping objects with different shapes, we use procedurally generated random geometric shapes in simulation~\cite{Bousmalis2017UsingSA}. Simulated images do not look realistic (see the left most images in Figure~\ref{fig:kuka-gans}) and models trained purely in simulation perform very poorly on real robots, making this an ideal task to evaluate simulation-to-real methods. Evaluations are performed using 4 robots, each with a set of 6 unseen objects. Each robot performs 102 grasps and drops any successfully grasped object back in the bin. Grasp success is reported as a percent average over all grasps.

\subsection{Robot 2 Setup}

\begin{figure}[ht]
\centering
\vspace{-5mm}
\subfloat[\footnotesize{\textbf{Single-bin grasping evaluation setup:} All robots are placed in front of the center bin, which contains all the 6 objects.} \label{fig:robot_2_eval_setup_single_bin}]{%
    \includegraphics[width=\linewidth]{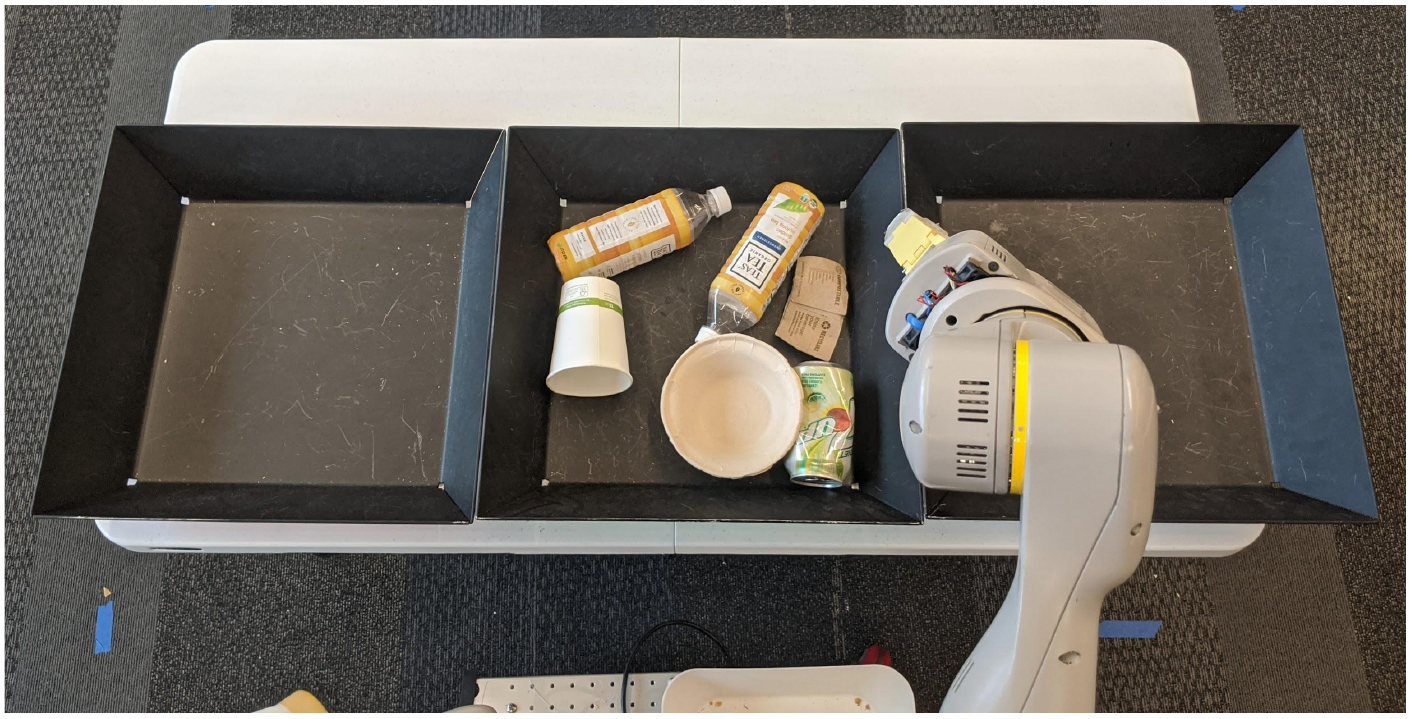}
}\newline
\subfloat[\footnotesize{\textbf{Multi-bin grasping evaluation setup:} some robots are centered with bins, \textbf{(i)} with the left bin, \textbf{(iii)} \& \textbf{(iv)} with the center bin, and \textbf{(vi)} with the right bin, while \textbf{(v)} and \textbf{(ii)} are off-set. Each setup contains 6 objects.} \label{fig:robot_2_eval_setup_multi_bin}]{%
    \includegraphics[width=\linewidth]{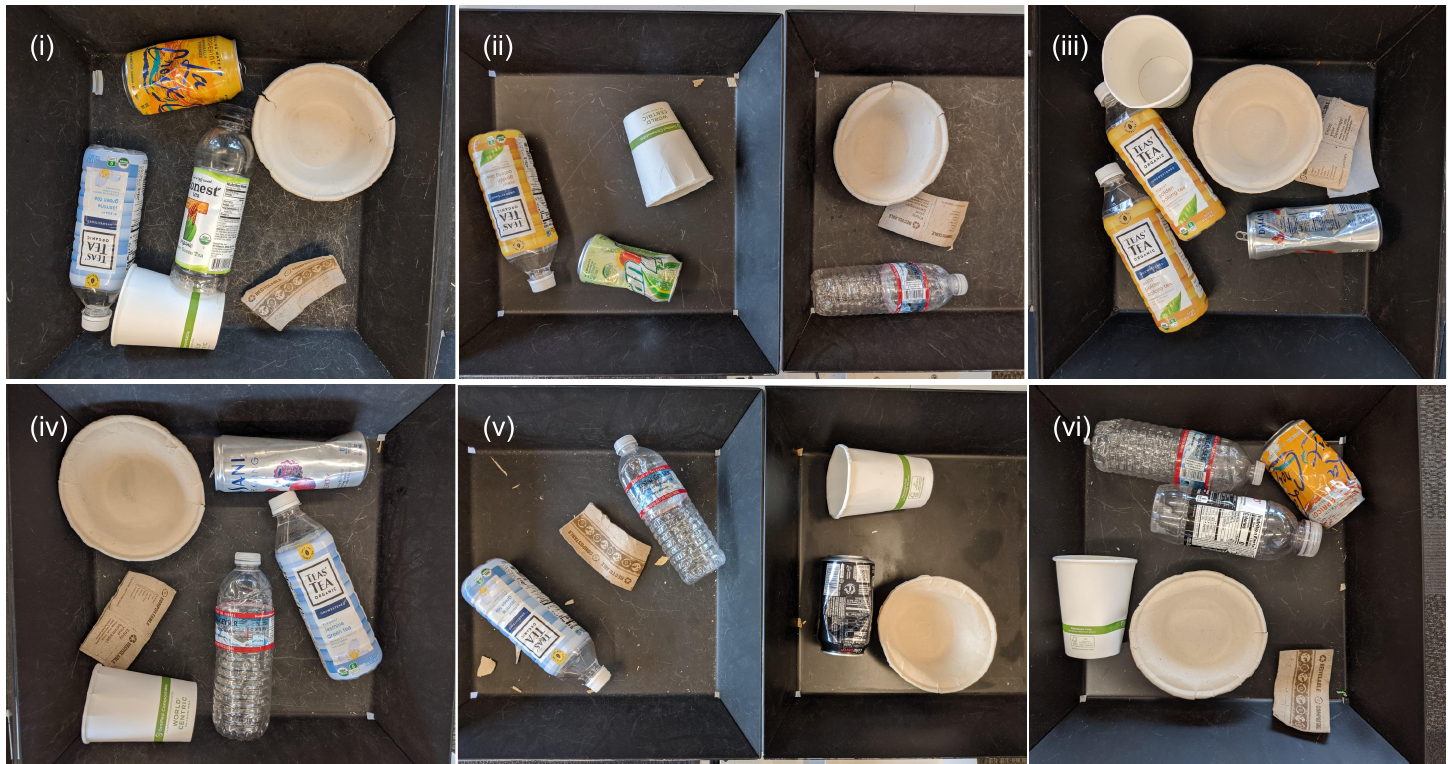}
}
\vspace{1mm}
\caption{\footnotesize Evaluation setup on Robot 2.}
\vspace{-2mm}
\label{fig:robot_2_eval_setup}
\end{figure}
We use robots to grasp trash-like items from three adjacent bins. In order to grasp from all areas of the three bins, the robot arm is mounted on a mobile base. The base is not controlled by the policy and remains fixed for the entire grasping episode, but is randomly positioned at the start of each episode. A learned policy must generalize to grasping from all three bins with a variety of camera angles. Real robot episodes are collected by using a scripted policy where the robot randomly drives to a location within the work-space in front of the three bins. A simulator is also built for this robot setup, and a large simulation-to-real visual gap (see Figure~\ref{fig:rl-cyclegan-diagram}) results in poor real world performance when models are trained without adaptation. 

We consider two types of evaluations, shown in Figure \ref{fig:robot_2_eval_setup_multi_bin}.
\textbf{Single-bin grasping:} robots are each placed in front of the center bin, which contains 6 objects (see Figure \ref{fig:robot_2_eval_setup_single_bin}). This evaluates grasping performance from a single base position form a single bin.
\textbf{Multi-bin grasping:} to evalaute grasping from all bins with varied base locations, robots are placed with some offsets with respect to the bins with objects also placed in different bins. In both types of evaluations, 6 robots are allowed 6 grasps and successfully grasped objects are placed outside the bin. This procedure is repeated 3 times for a total of 108 grasps, grasp success is reported as a percent average. 

\section{Experiments}
\label{eval}
We evaluate simulation-to-real methods for robotic grasping in a scenario where off-policy real-world data is available but may be limited, along with relatively cheap simulated experience. Learning from entirely off-policy real world data by itself is known to result in worse performance than on-policy fine-tuning~\cite{Kalashnikov2018ScalableDR}. The aim of our experiment is to understand whether RL-CycleGAN can bridge this gap in performance by utilizing simulated experience, and whether it can further reduce the amount of real-world data needed for good performance. We also compare RL-CycleGAN with state-of-the-art simulation-to-real methods for the robotic grasping tasks. Performance is evaluated in terms of the grasp success rate on the two robotic grasping systems described in the preceding section.

\rlc was evaluated across three sets of experiments.
In the first set, we trained various GAN approaches, then trained RL grasping models with simulations alone, but with the GAN applied to the simulated images. This investigates how well they address the visual simulation-to-reality gap for the grasping task.
In the second set of experiments, we reuse the real off-policy data used to train \rlc to also train the grasping model, mixing it with GAN adapted on-policy simulated data. We compare the improvements from including \rlc with varying amounts of real data. In the final experiments, we further fine-tune grasping models on-policy with real robots, while still using additional GAN-adapted on-policy simulated data.
Since on robot training is available,
in these final experiments
we restrict \rlc training to use very limited amounts of off-policy real data. 

\subsection{GANs For RL}
\label{gans_for_rl}

\begin{table}[t]
\caption{\footnotesize Robot 1 grasping performance for various models trained using simulations. First two are models trained with and without visual randomization applied in the simulator. The next four models utilize various GANs to adapt the simulated image to look more realistic, all GANs are trained with 580,000 real episodes.}
\vspace{-2mm}
\label{gan-compare}
\begin{center}
\begin{tabular}{lc}
\multicolumn{1}{l}{\bf Simulation-to-Real Model}   &\multicolumn{1}{c}{\bf Robot 1 Grasp Success} \\ \hline 
Sim-Only~\cite{James2018SimtoRealVS}            & 21\%\\
Randomized Sim~\cite{James2018SimtoRealVS}            & 37\%\\ \hline 
GAN            & 29\%\\
CycleGAN        & 61\%\\
GraspGAN        & 63\%\\
\rlc   & {\bf 70\%}\\
\end{tabular}
\end{center}
\vspace{-8mm}
\end{table}


We first establish a baseline for simulation-to-real world transfer without any real data or domain adaptation. As shown in Table~\ref{gan-compare}, our standard simulator, without any adaptation, results in a policy that only achieves 21\% grasp success in the real world, though the simulation success rate is 95\%. This indicates a large simulation-to-real gap. Incorporating randomization into the visual appearance of the arm, objects, bin and backgrounds~\cite{James2018SimtoRealVS} increases this performance to 37\%, but a large gap still remains. 

We next compare different GAN-based adaptation models, including ablations of our method. To evaluate the usefulness of the GAN for RL we train grasping models using the simulator only, but with the pre-trained GANs applied to the simulated images. Examples and a qualitative discussion of the various models is presented in Figure~\ref{fig:kuka-gans}. Table~\ref{gan-compare} shows the improved performance from using a CycleGAN, 61\% success, versus a regular GAN, 29\% success. The cycle consistency encourages retaining the position of objects and the arm, however, occasional objection deletion and addition is still observed.
The GraspGAN performs comparably (63\% success) to the CycleGAN, because of a grasping specific masking loss which avoids object deletion or addition, but overall image quality is less realistic especially with the robot arm. One hypothesis is that the domain-adversarial losses used by GraspGAN may restrict the realism of generated images, however, we do not test that here.
The grasping model trained with the \rlc performs the best (70\% success). The \rlc preserves task-salient information and produces realistic images, and does so with a general-purpose consistency loss that is based directly on the similarity of Q-values, without requiring manual identification of task-salient properties (e.g., object geometry).

\subsection{Mixing Real Data And Simulation}
\label{kuka_sim2real}

We investigate how RL models may be trained by mixing real off-policy data and a simulator with simulation-to-real adaptation via \rlcnospace. In this experiment, we measure how performance scales with the amount of real data.
First, an \rlc model is trained with the available real data as in the preceding section. The same real data is then reused during training of the final RL model. In this way, the RL process benefits from real off-policy data and {\it realistic} on-policy data generated after applying the \rlc to the simulated data. For baselines, we train grasping models with only the real off-policy data for Robot 1, and with a mix of real off-policy and simulated on-policy data for Robot 2.

For Robot 1, Table~\ref{kuka-table-1} shows significant improvement from \rlcnospace: using only 5,000 real-world trials for training the GAN and for the RL process improves the grasp success rate from 15\% to 75\%. It is important to note here that the real data is used in \emph{two} ways: to train the GAN and for RL.
Even with a large dataset of 580,000 real-world trials, we see significant improvements with the \rlcnospace, going from 87\% to 94\% grasp success. This is comparable to the state-of-the-art performance (96\%) described by~\cite{Kalashnikov2018ScalableDR}, which required lengthy on-robot fine-tuning. \rlc is able to achieve this performance with only previously collected off-policy trials, making for a much easier and more practical training procedure.

Simulation-to-real transfer via the \rlc sees similar significant improvements for Robot 2 in Table~\ref{sorty-table-1}. With only 3,000 real episodes, we see a performance improvement from 13\% to 72\%. With 80,000 real episodes, the model trained with \rlc reaches state-of-the-art performance at 95\% grasp success. Similar performance is seen for multi-bin grasping with randomized base locations
at 93\% grasp success, showing that \rlc generalizes well to different grasp locations and camera angles.

\begin{table}[t]
\caption{Grasping success for Robot 1 with varying amounts of real data and the corresponding improvements from including simulations with simulation-to-real methods: GraspGAN and \rlcnospace.}
\vspace{-2mm}
\label{kuka-table-1}
\begin{center}
\begin{tabular}{lcccc}
\multicolumn{1}{c}{\bf Episodes}  &\multicolumn{3}{c}{\bf Robot 1 Grasp Success} \\
& Real Only & GraspGAN & \rlc \\
 \hline 
5,000 &  15\% & - & 75\%\\
28,000 &  16\% & -& 86\%\\
580,000 &  87\% & 89\%& {\bf 94\%}  \\
\end{tabular}
\end{center}
\vspace{-2mm}
\end{table}

\begin{table}[t]
\caption{Grasping success with Robot 2 setup using simulation and varying amounts of real episodes versus models that use \rlc to adapt the simulated images.}
\vspace{-2mm}
\label{sorty-table-1}
\begin{center}
\begin{tabular}{lccc}
\multicolumn{1}{c}{\bf Off-policy episodes}  &\multicolumn{2}{c}{\bf Robot 2 Grasp Success} \\
& Sim+Real & \rlc \\
\hline 
\multicolumn{3}{c}{\bf Single-bin grasping, centered}  \\
3,000 &  13\% & 72\% \\
5,000 &  12\% & 76\% \\
10,000 &  10\% & 84\% \\
80,000 &  36\% & {\bf 95\%}\\ 
\hline 
\multicolumn{3}{c}{\bf Multi-bin grasping, randomized location}  \\
80,000 &  33\% & {\bf 93\%} \\
\end{tabular}
\end{center}
\vspace{-8mm}
\end{table}

\subsection{On-robot Fine-tuning}
Grasping models trained as described in the preceding section can be further fine-tuned with on-robot training. During fine-tuning, real on-policy data from the robot is mixed with on-policy simulated data adapted with \rlcnospace. To compare with simulation-to-real methods such as RCAN~\cite{James2018SimtoRealVS}, which only uses on-policy real data and no off-policy data, we restrict the amount of off-policy real data used to train \rlc to 5,000 grasps, about two orders of magnitude less than required for state-of-the-art methods trained on real data~\cite{Kalashnikov2018ScalableDR}. In the absence of real data, RCAN allows for zero-shot transfer to the real world at 70\% grasp success, which significantly outperforms randomization alone. However, real on-policy fine-tuning for 28,000 episodes was required for RCAN to reach 94\% grasp success. We find that \rlc can be reliably trained with only a few thousand episodes. With 5,000 off-policy episodes \rlc achieves 75\% grasp success, which when fine-tuned over 10,000 on-policy episodes achieves the same performance as RCAN at 94\% (Table~\ref{rcan-comp}).

\begin{table}
\caption{Grasping success on Robot 1 for \rlc and RCAN with on-policy fine-tuning. While RCAN achieves decent performance with zero real data RL-CycleGAN does not require domain randomization and after on-policy training performs similarly to RCAN.}
\label{rcan-comp}
\vspace{-4mm}
\begin{center}
\begin{tabular}{p{2.15cm}p{0.8cm}p{0.8cm}p{0.8cm}p{1.2cm}}
 &\multicolumn{2}{c}{\bf Episodes}  &\multicolumn{1}{c}{\bf Domain} &\multicolumn{1}{c}{\bf Robot 1 } \\
{\bf Model}& Off-policy & On-policy & {\bf Rand.}& {\bf Grasp Success}
\\ \hline 
RCAN~\cite{James2018SimtoRealVS}    & -     & -      & \hspace{0.3cm}\cmark & \hspace{0.25cm}70\%\\
RCAN~\cite{James2018SimtoRealVS}    & -     & 5,000 & \hspace{0.3cm}\cmark & \hspace{0.25cm}91\%\\
\rlc                                & 5,000 & 5,000  & \hspace{0.3cm}\xmark  & \hspace{0.25cm}90\%\\
RCAN~\cite{James2018SimtoRealVS}    & -     & 28,000 & \hspace{0.3cm}\cmark & \hspace{0.25cm}{\bf 94}\%\\
\rlc                                &5,000 & 10,000  & \hspace{0.3cm}\xmark   & \hspace{0.25cm}{\bf 94}\%\\
\end{tabular}
\end{center}
\vspace{-8mm}
\end{table}

\section{Conclusion}
\label{conclusion}
We have presented the \rlc to address the visual simulation-to-real gap, and showed it significantly improves real world vision-based robotics with two varied grasping setups. Incorporating an RL scene consistency loss along with the CycleGAN losses provides a natural separation of the {\it style}, which may be adapted to look more realistic, and the relevant {\it semantics} for RL that must be preserved. This removes the need for task-specific feature engineering, such as generating scene segmentation masks for GraspGAN or defining a canonical scene and simulation randomization for RCAN. 

\rlc only addresses the visual gap, and not any physics based simulation-to-real differences. Handling these cases requires extending the \rlc to adapt the entire state-action trajectory of an episode instead of a single state image, which is left for future work. The GANs presented in this work are deterministic, since no random noise is sampled and the same input simulated image is always adapted to the same realistic version. \rlc may be extended to produce stochastic outputs by incorporating ideas from recent works like Augmented-CycleGAN~\cite{Almahairi2018} or BicycleGAN~\cite{Zhu2017UnpairedIT}

For both robotic grasping setups we see large performance gains
by incorporating on-policy simulator data adapted with \rlc.
With Robot 1, we require 20 times fewer real grasps (28,000) with \rlc to attain the performance from using 580,000 real grasps. When using all 580,000 real grasps, \rlc (94\% success) is comparable to the state-of-the-art (96\%) but without requiring costly on-robot training. We see even larger improvements with Robot 2, where the \rlc achieves 72\% grasp success at centered single-bin grasping, with only 3,000 real grasps, vastly outperforming the 36\% grasp success from a baseline model trained with 80,000 grasps. With 80,000 grasps, the \rlc trained model achieves state-of-the-art results, with 95\% success for centered single-bin grasping and 93\% success for multi-bin grasping from randomized locations.
\section{Acknowledgements}
We would like to thank Ivonne Fajardo, Noah Brown and Benjamin Swanson for overseeing the robot operations, Paul Wohlhart and Konstantinos Bousmalis for valuable discussions on image generation, Anthony Brohan and Yao Lu for help with training infrastructure and 
Chad Richards, Vincent Vanhoucke and Mrinal Kalakrishnan for comments on the paper.

{\small
\bibliographystyle{ieee_fullname}
\bibliography{main}
}

\clearpage
\section*{Appendix}
\setcounter{section}{0}
\renewcommand\thesection{\Alph{section}}

\section{Model Training Details}
\label{app-model-training}

\begin{figure}[h]
\begin{center}
\includegraphics[width=0.8\linewidth]{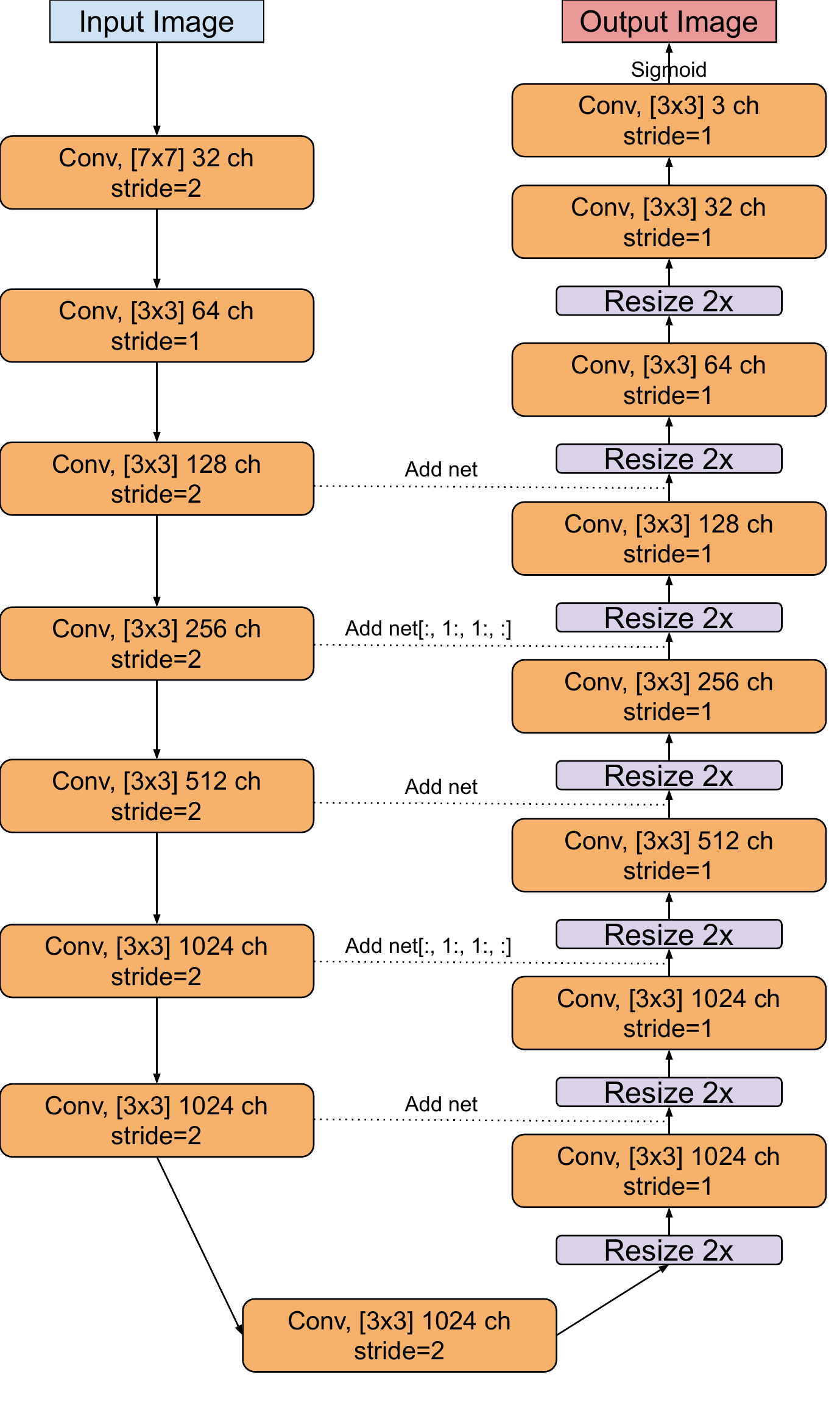}
\end{center}
\caption{The U-net architecture used for $G$ and $F$ generator networks. Each convolutional block is shown with kernel size, number of filters and stride. Spectral normalization is applied to all convolutions. Images are resized to be twice as high and wide using nearest neighbor interpolation. Intermediate outputs from the down-convolutions (left) are added during the up-convolutions (right) as shown by the dotted lines, in two cases the first row and column are dropped during addition to match sizes. Instance normalization~\cite{Ulyanov2016InstanceNT} is applied to all convolutions except the final output convolution. }
\label{fig:unet}
\end{figure}

We use data-sets consisting of grasping episodes from simulation and real robots. For Robot 1 and \sorty the data-sets are 580,000 and 80,000 real robot episodes respectively. Both data-sets are collected by starting with a human-designed scripted policy, which succeeds a small fraction of the time. Models are trained with this data, and periodically, those models are deployed to the robot to collect data from a better policy. When collecting data, random exploration noise is added to collect more diverse data. For this paper, we randomly subsample smaller datasets from these larger sets, to study the performance when using varying amounts of real episodes.
For both setups several million simulated episodes are also generated during on-policy training.

Typical episodes contain 6-10 states represented by a $512$ pixel high, $640$ pixel wide RGB image. To increase data diversity images are randomly cropped to $472$x$472$ during training. At inference time, the center $472$x$472$ square from the image is used.
The generator for the GAN is a convolutional neural network with a U-Net architecture~\cite{Ronneberger2015UNetCN} as shown in ~\ref{fig:unet}. The discriminator is smaller convolutional neural network that operates on three scales of the input image. Both networks are described in detail in~\cite{Bousmalis2017UsingSA}. The robotic grasping task is trained via QT-Opt with the Q-function represented as a convolutional neural network (see~\cite{Kalashnikov2018ScalableDR} for architecture). \rlc jointly trains the Cycle-GAN along with the Q-function during QT-Opt. Models are trained on Google TPUv3 Pod as in~\cite{Brock2018LargeSG} and required $bfloat16$ precision training to fit in memory. Each batch had 8 real images and 8 simulated images. We use Adam optimizer~\cite{Kingma2014AdamAM} with $\beta_1$ = 0.1 and $\beta_2$ = 0.999 and a constant learning rate of 0.0001. We employ Spectral Normalization~\cite{Zhang2018SelfAttentionGA} in the GAN generator networks and find that it improves stability.

\begin{figure}[h]
\begin{center}
\includegraphics[width=\linewidth]{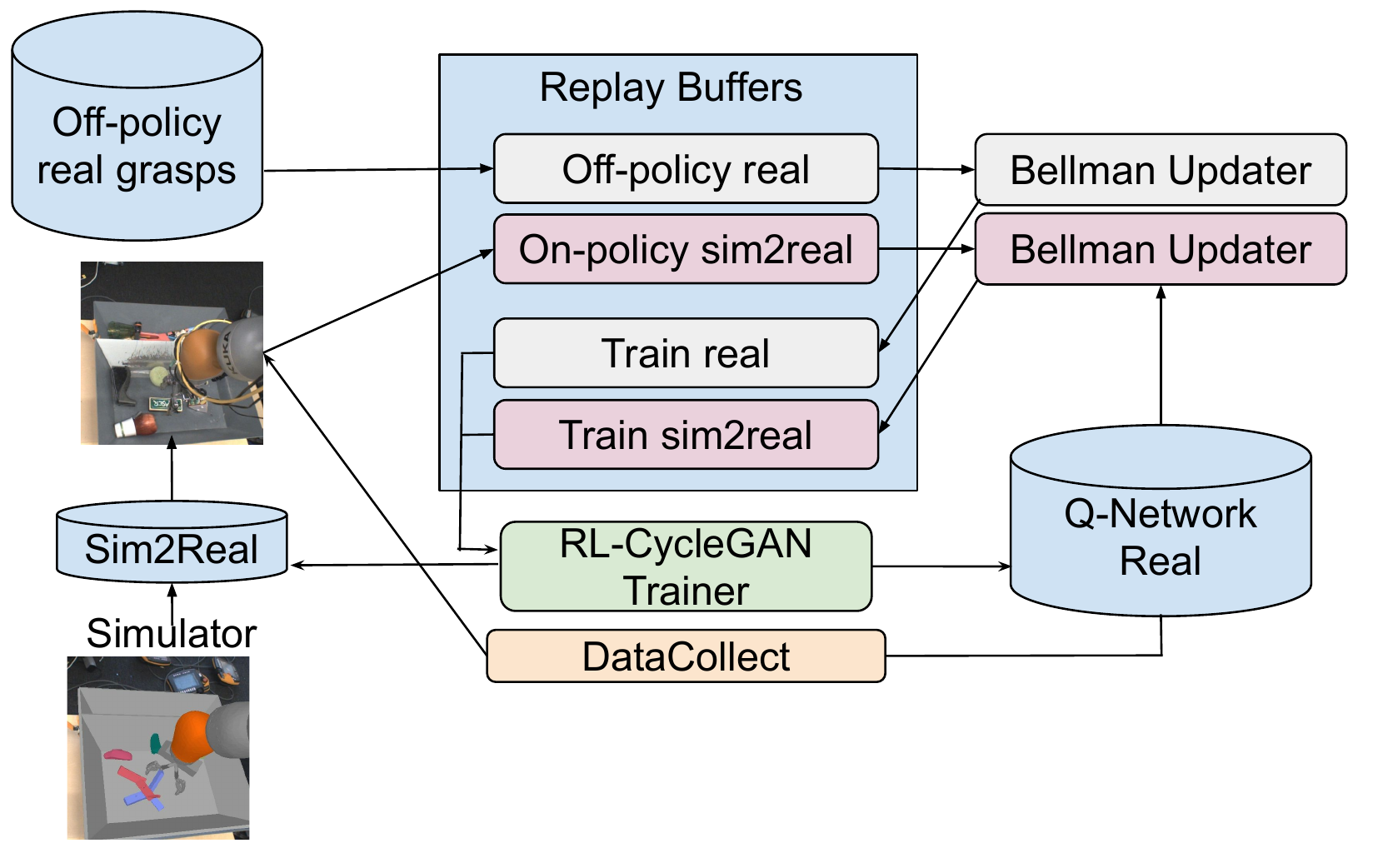}
\end{center}
\caption{Training \rlc via QT-Opt.}
\label{fig:qtopt_sim2real}
\end{figure}

Figure~\ref{fig:qtopt_sim2real} shows how we train \rlc via QT-Opt. Images from a simulator are transformed by Sim2Real generator $G$ and then passed to $Q_{real}$ to generate an action. In this way, on-policy (w.r.t $Q_{real}$) episodes are generated in the simulation-to-real environment. Off-policy real grasps are read from disk. Separate replay buffers and bellman update instances are used for the off-policy real and off-policy simulation-to-real data. \rlc is trained with batches with equal parts from real and simulation-to-real data. During training we evaluate the performance of both $Q_{sim}$ and $Q_{real}$ in the simulator, with simulation-to-real applied prior to evaluating $Q_{real}$.
Training converges when certain conditions are met, the simulation-to-real images look realistic, the cycled images look reasonable with a reasonably low $\mathcal{L}_{cyc}$, both $Q_{sim}$ and $Q_{real}$ perform well along with a reasonably low $\mathcal{L}_{RL-scene}$. A final $Q_{real}$ is trained from scratch with the pre-trained and fixed Sim2Real generator $G$. This phase of training is as before, but with only the reinforcement learning loss.

\begin{table}[t]
\caption{The various losses, their relative weights and the networks they affect for \rlc.}
\label{rlc-deets}
\begin{center}
\begin{tabular}{lcc}
{\bf Loss}& {\bf Relative Weight ($\lambda$)} &{\bf Networks Updated} \\
 \hline 
$\mathcal{L}_{GAN}$& 1 & $G$, $F$, $D_X$, $D_Y$\\
$\mathcal{L}_{cycle}$& 10 & $G$, $F$\\
$\mathcal{L}_{RL}$& 10 & $Q_{sim}$, $Q_{real}$\\
$\mathcal{L}_{RL-scene}$& 10 & $G$, $F$\\
\end{tabular}
\end{center}
\end{table}

Depending on the relative loss weights, $ \lambda $, \rlc might experience a particular mode of collapse, where $Q$ outputs incorrect, uniform Q-values that give a spuriously low $\mathcal{L}_{RL-scene}$. This mode collapse can be caught by monitoring performance of $Q$ during training and tuning $\lambda$ appropriately. \rlc involves multiple losses which are selectively applied to the various neural network components. The relative loss weights and the neural networks affected by the various losses is listed in Table~\ref{rlc-deets}.

We found that although adding a $Q$-network to CycleGAN improved performance, it was critical to maintain some separation between the two during optimization time.
When $\mathcal{L}_{RL}$ and $\mathcal{L}_{RL-scene}$ were optimized entirely end-to-end, saliency analysis showed the $Q$-value for generated images was mostly dependent on generators $G$ and $F$, rather than $Q$. We theorized the generators were computing the $Q$-value needed to minimize $\mathcal{L}_{RL-scene}$, embedding them within the generated image, and the $Q$-networks were simply decoding the embedded value. Such a $Q$-network generalizes poorly, does not understand the scene, and consequently does not provide a useful RL-scene consistency loss.

To fix this, the gradient for $\mathcal{L}_{RL}$ is only applied to $Q$, and the gradient for $\mathcal{L}_{RL-scene}$ is only applied to $G$ and $F$. Note that in both cases, we still compute the full backward pass (there is no stop gradient), but we selectively choose which networks the gradient is applied to. Doing so makes it harder for the optimization to learn the poor encoding-decoding behavior mentioned above.

\begin{table}[t]
\caption{The impact of using $\lambda_{RL-real}$. With only 3,000 real episodes a small real loss weight of $0.1$ is optimal while with a large data-set of 80,000 real episodes a real loss weight of $2.0$ was found to be best.}
\label{real_loss_weight}
\begin{center}
\begin{tabular}{lccc}
\multicolumn{1}{c}{\bf Off-policy}  &\multicolumn{1}{c}{ $\lambda_{RL-real}$} &\multicolumn{1}{c}{\bf Robot 2 } \\
\multicolumn{1}{c}{\bf episodes}  &\multicolumn{1}{c}{} &\multicolumn{1}{c}{\bf Grasp Success} \\
\hline 
3,000 &  1.0 & 66\% \\
3,000 &  0.1 & 72\% \\
80,000 &  1.0 & 91\%\\ 
80,000 &  2.0 &  95\%\\ 
\hline 
\end{tabular}
\end{center}
\end{table}

Since we train with batches of equal amounts of data from real data and from the simulator, we weight the loss from the real data depending on how many real episodes are available. While training the final $Q_{real}$ a weighting term $ \lambda_{RL-real}$ is applied,

 $\mathcal{L}_{RL} = \sum_{Sim2Real} \mathcal{L}_{RL} + \lambda_{RL-real} \sum_{Real}  \mathcal{L}_{RL}$

For all experiments we use $\lambda_{RL-real}=0.1$ if using 10,000 real episodes or fewer, and $\lambda_{RL-real}=2$ with 80,000 real episodes or more. Ablation results are shown in Table~\ref{real_loss_weight}.


\section{Robot Simulated Objects}
\label{app-robot-2}

\begin{figure}[h]
\begin{center}
\includegraphics[width=0.8\linewidth]{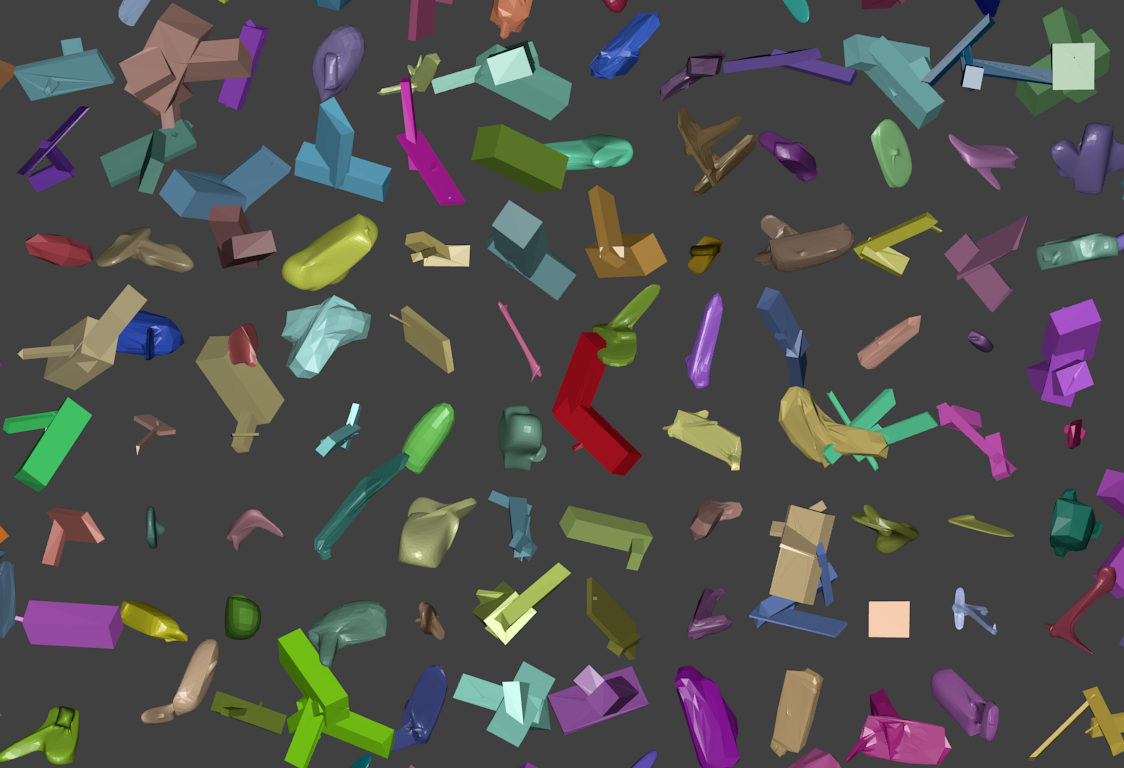}
\includegraphics[width=0.8\linewidth]{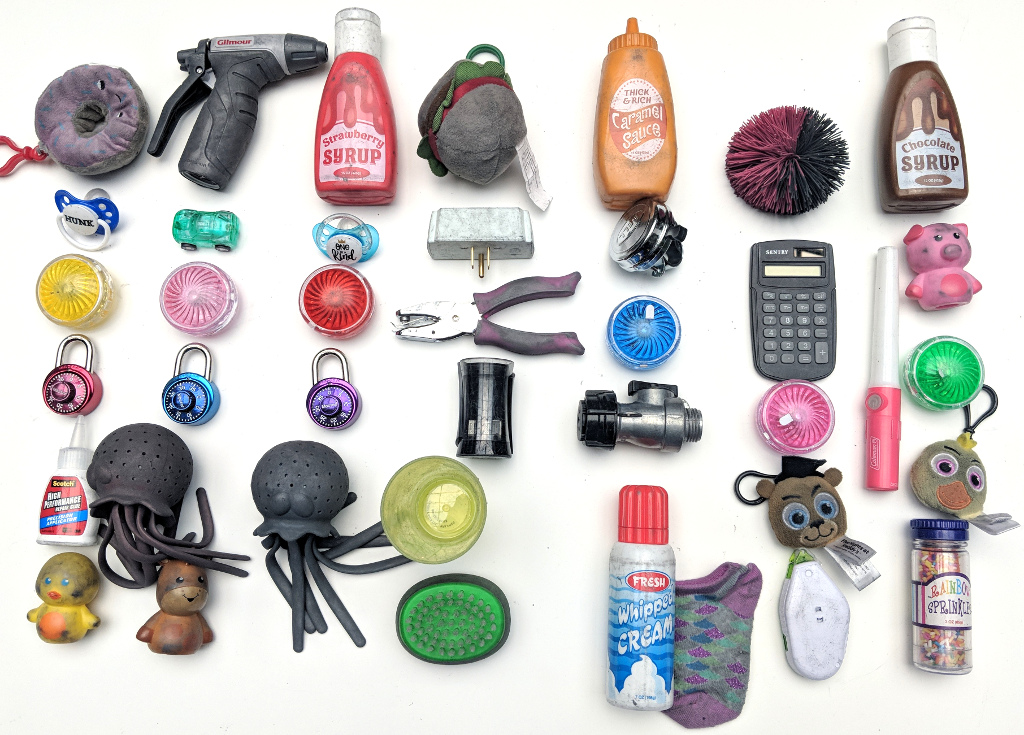}
\end{center}
\caption{Robot 1: procedural objects generated in the simulator (top) and the unseen objects using during evaluation (bottom).}
\label{fig:robot_1_sim_dataset}
\end{figure}

\begin{figure}[h]
\begin{center}
\includegraphics[width=0.8\linewidth]{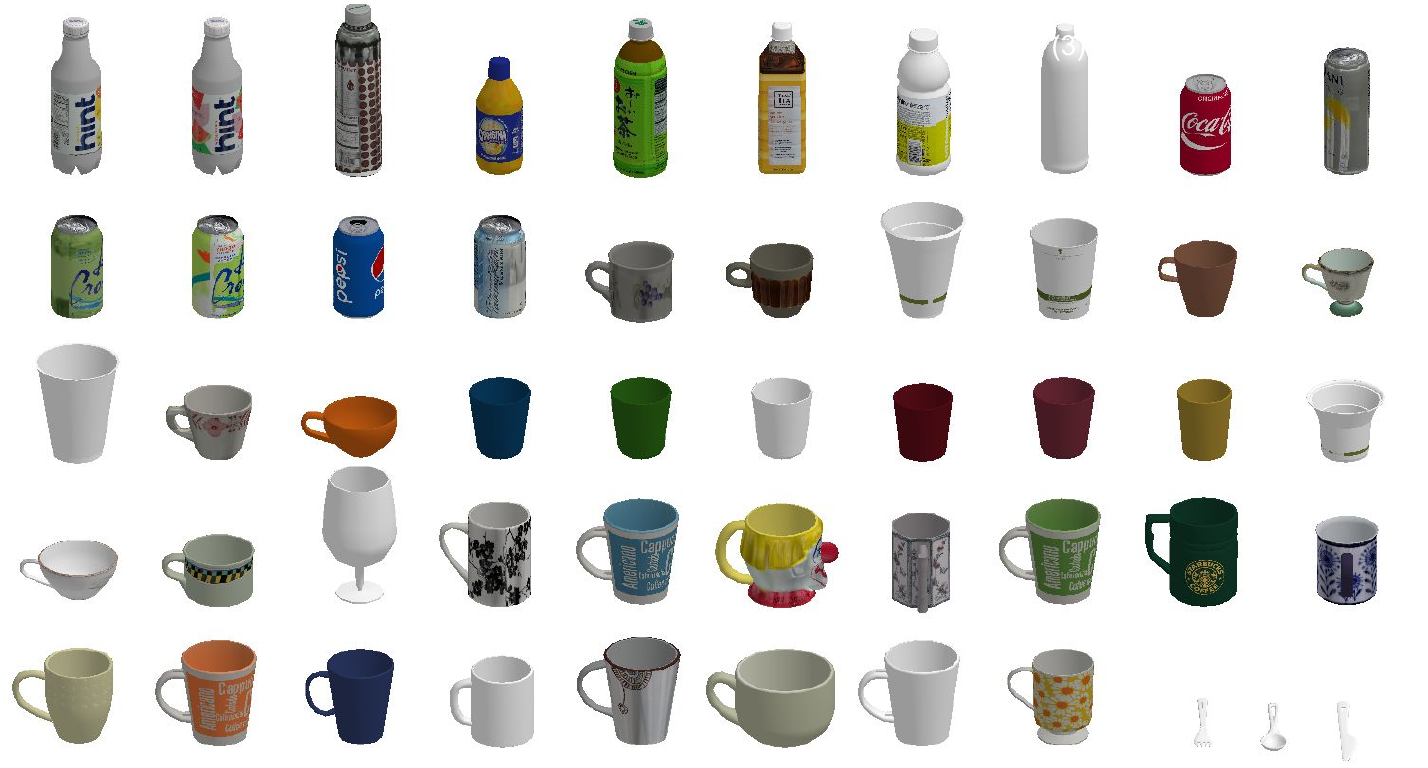}
\end{center}
\caption{Robot 2: simulated versions of 51 common real world objects are used when training the \rlc, including plastic bottles (8), coffee cups (18), plastic utensils (3), drink cans (6), mugs (15), and wine glass (1).}
\label{fig:robot_2_sim_dataset}
\end{figure}

The goal with Robot 1 to be able to grasp unseen objects during evaluation. In simulation we procedurally generate objects with random shapes by attaching rectangular prisms at random locations and orientations. These procedural objects and the actual unseen objects used during evaluation are shown in Figure~\ref{fig:robot_1_sim_dataset}.

Since Robot 2 grasps trash-like objects we mimic the simulated objects more closely with the real world object. We create simulated versions of 51 common real world objects are created: plastic bottles (8), coffee cups (18), plastic utensils (3), drink cans (6), mugs (15), and wine glass (1), shown in Figure~\ref{fig:robot_2_sim_dataset}. These do not cover all the real world objects used evaluation.

\end{document}